\documentclass[11pt,a4paper]{article}
\usepackage[hyperref]{acl2019}
\usepackage{times}
\usepackage{latexsym}

\usepackage{url}

\usepackage{graphicx}
\usepackage{amsmath}
\usepackage{amsfonts}
\usepackage[ruled, lined, vlined]{algorithm2e}
\usepackage{multirow}
\usepackage{subfig}
\aclfinalcopy 
\def\aclpaperid{112} 

\title{Towards End-to-End Learning for Efficient Dialogue Agent by Modeling Looking-ahead Ability}

\author{Zhuoxuan Jiang\textsuperscript{1}, Xian-Ling Mao\textsuperscript{2}\thanks{~~Xian-Ling Mao is the corresponding author.}, Ziming Huang\textsuperscript{3}, Jie Ma\textsuperscript{3}, Shaochun Li\textsuperscript{3} \\
  \textsuperscript{1}IBM Research / Shanghai, China \\
  \textsuperscript{2}Beijing Institute of Technology / Beijing, China \\
  \textsuperscript{3}IBM Research / Beijing, China \\
  \texttt{\{jzxjiang,hzmzi,bjmajie,lishaoc\}@cn.ibm.com, maoxl@bit.edu.cn} \\
}

\date{}

\begin{document}
\maketitle
\begin{abstract}

Learning an efficient manager of dialogue agent from data with little manual intervention is important, especially for goal-oriented dialogues. However, existing methods either take too many manual efforts (e.g. reinforcement learning methods) or cannot guarantee the dialogue efficiency (e.g. sequence-to-sequence methods). In this paper, we address this problem by proposing a novel end-to-end learning model to train a dialogue agent that can look ahead for several future turns and generate an optimal response to make the dialogue efficient. Our method is data-driven and does not require too much manual work for intervention during system design. We evaluate our method on two datasets of different scenarios and the experimental results demonstrate the efficiency of our model.
\end{abstract}

\section{Introduction}


Research for dialogue system attracts a lot of attentions recently due to its potential huge value of reducing human cost in many commercial domains, such as restaurant reservation~\cite{dstc3}, travel planning~\cite{hierarchical} and retail service~\cite{retail}.
The majority of dialogue agents with goals are expected to be efficient to complete tasks with as few as possible dialogue turns, in contrast with those chit-chat counterparts~\cite{chat1}. The following two examples show the comparison of efficient and inefficient cases respectively. The senarios is restaurant reservation and the agent's goal is to reserve a table at noon.
\begin{quote}
	Efficient example:\\
	``\textit{Human}: We don’t have empty tables at 11 o’clock tomorrow. All are reserved.''
	``\textit{Agent}: How about 12 o'clock? We are also okay then.''
\end{quote}
\begin{quote}
	Inefficient example:\\
	``\textit{Human}: We don’t have empty tables at 11 o’clock tomorrow. All are reserved.''
	``\textit{Agent}: What else time is available?''
	``\textit{Human}: 12 o'clock is ok.''\\
	``\textit{Agent}: All right. We want that time.''
\end{quote}
For expressing the same opinion, the inefficient example consumes four turns while the efficient example only needs two. As it can be seen, the efficiency is important for goal-oriented dialogue systems to achieve goals in a rapid way.

Usually, a dialogue system consists of a pipeline of natural language understanding (NLU), dialogue management (DM) and natural language generation (NLG), where the DM part is treat as two separate components: dialogue state tracking (DST) and dialogue control (DC, i.e. dialogue policy selection). The DM part is widely considered to be relevant to the dialogue's efficiency, because it makes decisions on what to say for the next turn. Recently, methods based on reinforcement learning are proposed for the policy selection component to build efficient dialogue systems. However, there are some drawbacks of reinforcement learning based methods. For example, they requires lots of human work to design the learning strategy. Also a real-world environment which is essential for the agent to learn from is expensive, such as from domain experts. Moreover, training the dialogue manager as a two separate components could lead to error propagation issue~\cite{multitask1}.

In addition to reinforcement learning based methods, sequence-to-sequence based methods are also popular recently, because they can learn a dialogue agent purely from data and almost without too many human efforts. The  error propagation issue can also be reduced because they are end-to-end, and they have better scalability for different scenarios. However, it is difficult to build efficient dialogue agents by those methods since their objective functions for training models are usually inclined to general responses, such as \textit{I don't know}, \textit{yes} and \textit{OK}, or often generate the same response for totally different contexts because the contextual information is not well-modeled by those methods~\cite{evaluate2}.

In this paper, we address the problem of learning an efficient dialogue manager from the perspective of reducing manual intervention and error propagation, and propose a new sequence-to-sequence based approach. The proposed end-to-end model contains a novel looking-ahead module for dialogue manager to learn the looking-ahead ability. Our intuition is that by predicting the future several dialogue turns, the agent could make a better decision of what to say for current turn, and therefore goals could be sooner achieved in a long run.

More specifically, our model includes three modules: (1) encoding module, (2) looking-ahead module, and (3) decoding module. At each dialogue turn, three kinds of information, the goals, historical utterances and the current user utterance, are utilized.  First they are encoded by three separate Bidirectional Gated Recurrent Units (BiGRU) models. Then the three encoded embeddings are concatenated to one vector, which is then sent to a new bidirectional neural network that can look ahead for several turns. The decoding module will generate utterances for each turn through a learned  language model. At last, by considering all the predicted future utterances, a new real system utterance for the next turn is re-generated by using an attention model through the same language model.

Our proposed approach has several advantages. First, it is an end-to-end model and does not take too many human efforts for system design. Although the goals should be handcrafted for specific scenario, the number of goals is small and it is a relatively easy work. Moreover, compared with naive sequence-to-sequence based models, our agent can make the dialogue more efficient by modeling the looking-ahead ability. Experimental results show that our model performs better than baselines on two datasets from different domains, which could suggest that our model is also scalable to various domains.

The contributions in this paper include:

\begin{itemize}
  \item We identify the problem that how to make dialogues efficient by exploiting as little as possible manual intervention during system design from the perspective of end-to-end deep learning.
  \item We propose a novel end-to-end and data-driven model that enables the dialgoue agent to learn to look ahead and make efficient decisions of what to say for the next turn.
  \item Experiments conducted on two datasets demonstrate that our model performs better over baselines and can be applied to different domains.
\end{itemize}

\section{Related Work}

In most situations, the dialogue systems require handcrafted definition of dialogue states and dialogue policies~\cite{pomdp,challenge,neg2,jd}. Those methods make the pipeline of dialogue systems clear to design and easy to maintain, but suffer from the massive expensive human efforts and the error propagation issue~\cite{errorpro1,errorpro2}.


Reinforcement learning based methods for dialogue policy selection are widely studied recently~\cite{BBQ,movie,SIGDIAL,continue}. These methods only need human to design the learning strategies and do not require massive training data. However, the expensive domain knowledge and human expert efforts for agents to learn from are necessary~\cite{withhuman,bootstraping}. Therefore, hybrid methods that integrate supervised learning and reinforcement learning are proposed recently~\cite{hybrid,kefu}. Thus, collecting massive training data becomes another manual work.

More recently, end-to-end dialogue systems attract much attention because almost no human efforts are required and they are scalable for different domains~\cite{EACL,IJCNLP,deal,goal3}, especially with sequence-to-sequence based models~\cite{seq2seq}. Although those models have been proved to be effective on chit-chat conversations~\cite{chat1,chat2,chat3}, how to build agents that are goal-oriented with efficient dialogue managers  through end-to-end approaches still remains questionable~\cite{goal1,goal2}, and we investigate the question in this paper.

\begin{figure*}
  \centering
  \includegraphics[height=4.4in,width=6.3in]{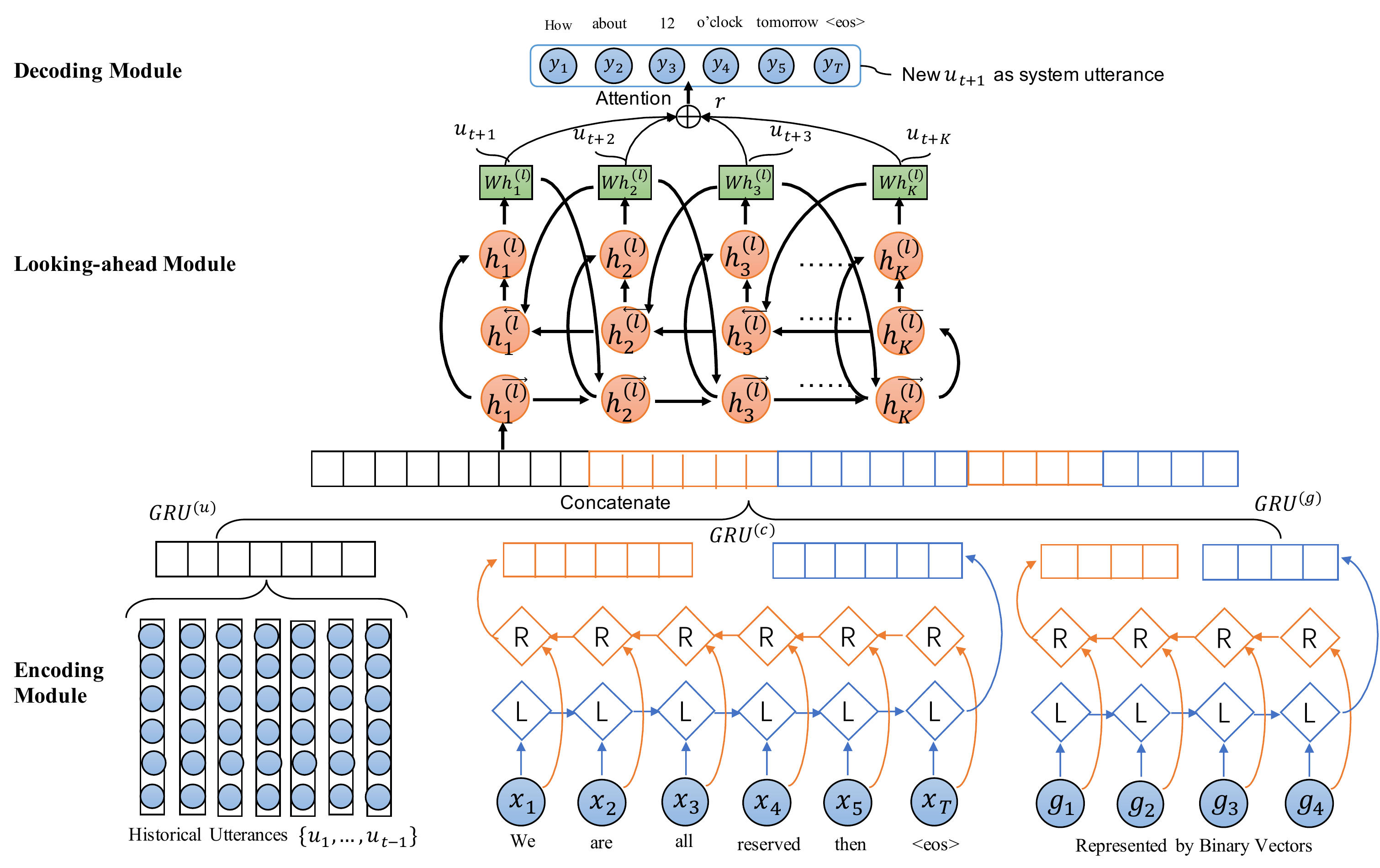}
  \caption{End-to-end model for learning looking-ahead ability.\label{fig:model}}
\end{figure*}

Our idea of enabling the agent to be efficient by modeling looking-ahead ability is inspired by the AI Planning concept, which is a traditional searching technology in the field of AI, and is suitable for goal-based tasks, such as robotics control~\cite{ai}. Recently, the concept is borrowed to dialogue system communities and integrated into deep learning models. For example, a trade-off method for training the agents neither with real human nor with user simulators is proposed, in order to obtain better policy learning results~\cite{DDQ}. In addition, at earlier time, the planning idea has been utilized for improving the dialogue generation task~\cite{plan1,plan2}.

\section{End-to-end Dialogue Model}

We propose an end-to-end model that contains three modules: (1) encoding module, (2) looking-ahead module, and (3) decoding module. Figure~\ref{fig:model} shows the model architecture. We leverage Bidirectional GRU models~\cite{gru1} to encode agent goals, historical and current utterances. Then the obtained representations by encoding goals and utterances are regarded as inputs of the looking-ahead module, and they are used to predict several future turns. At last the predicted future turns are merged by an attention model and the new real system utterance is generated for the next turn.

Suppose for each dialogue session we have $T$ turns, and we do not distinguish whether it is user's turn or system's turn. If the agent has $S$ goals that are denoted as $g=\{g_1,g_2,...,g_S\}$, each goal is formalized as a binary vector. For example in the restaurant reservation scenario, we can define that each variate in the vector $[1,0]$ corresponds to a yes-no condition, such as the $1$ means agent accepts bar table and the $0$ means agent does not want to change time. As to the utterance information, imagine at turn $t\in\{1,...,T\}$, we denote utterances $\{u_1,...,u_{t-1}\}\in \mathbb{U}$ for historical ones and $u_{t}\in \mathbb{U}$ for current user utterance. Our model predicts the system and user utterances $\{u_{t+1},u_{t+2},...,u_{t+K}\}$ for the next $K$ turns and then a new $u_{t+1}$ is generated as the system utterance after considering all the predicted turns. The model separates the current user utterance from historical ones in order to highlight the user's current states. In general, the model is end-to-end and needs little human intervention or domain knowledge.


\subsection{Encoding Module}

In this module, the agent goals, historical utterances within the dialogue session, and the current user utterance are encoded by using three GRU models which is expected to learn long-range temporal dependencies~\cite{gru}. $GRU^{(g)}$ is defined to encode agent's goals $g$ and the final hidden state $h^{(g)}$ is taken as the representation of goals. The input of $GRU^{(g)}$ is a one-hot binary vector with length $S$. 
$GRU^{(u)}$ is used to encode the historical utterances, and $GRU^{(c)}$ is used to encode the current user utterance. $h^{(u)}$ and $h^{(c)}$ are denoted as the final encoded representations of $GRU^{(u)}$ and $GRU^{(c)}$ respectively.

To get the $i$-th hidden state for the three GRUs, respective inputs include the previous hidden state $h_{i-1}^{(g)}$, $h_{i-1}^{(u)}$ or $h_{i-1}^{(c)}$, and the embeddings of current observations, $E(g_i)$, $E(u_i)$ or $E(x_i)$, where $g_i$ is a goal, $u_i$ is an utterance and $x_i$ is a token. For the textual tokens, we use the Word2vec embeddings as their representations~\cite{word2vec}. Then the token embeddings are averaged to represent utterances. The formal denotation of the hidden states for the three GRU models is:

\begin{equation}
h_i^{(g)}=GRU^{(g)}(h_{i-1}^{(g)}, E(g_{i})),
\end{equation}
\begin{equation}
h_i^{(u)}=GRU^{(u)}(h_{i-1}^{(u)}, E(u_{i})),
\end{equation}
\begin{equation}
h_i^{(c)}=GRU^{(c)}(h_{i-1}^{(c)}, E(x_{i})),
\end{equation}
where $E(\cdot)$ represents the embeddings.

The final output of the encoding module is a concatenation of $h^{(g)}$, $h^{(u)}$ and $h^{(c)}$, which is denoted as $h_1^{\overrightarrow{l}}=[h^{(g)},h^{(u)},h^{(c)}]$. $h_1^{\overrightarrow{l}}$ serves as the input of the following looking-ahead module. The right arrow means the initial direction to train the looking-ahead module is from the current to the future.

\subsection{Looking-ahead Module}

With the input of $h_1^{\overrightarrow{l}}$, this module predicts several future dialogue turns. Since the process is sequential, we propose a recurrent  neural network to model the process. In order to exploit the predicted information for later generating a real system utterance, another recurrent neural network is used to backtrack the information from future to current. To reduce the computing cost, the two neural networks share the same parameters, and the whole looking-ahead module looks similar to a bidirectional GRU as shown in Figure~\ref{fig:model}.

We denote the module as $GRU^{(l)}$. $\{h_k^{(l)}|k>0\}$ represent the predicted hidden states for future turns. To get $h_k^{(l)}$, the hidden states from two directions, $h_k^{\overrightarrow{l}}$ and $h_k^{\overleftarrow{l}}$, are concatenated. To calculate each $h_k^{\overrightarrow{l}}$ or $h_k^{\overleftarrow{l}}$, their inputs include the previous hidden state and the previously-predicted hidden state. Formally, suppose we look ahead for $K$ turns, the hidden state of $h_k^{(l)}$ is calculated as following:
\begin{equation}
h_k^{\overrightarrow{l}}=GRU^{\overrightarrow{l}}(h_{k-1}^{\overrightarrow{l}},Wh_{k-1}^{(l)}),
\end{equation}
\begin{equation}
h_k^{\overleftarrow{l}}=GRU^{\overleftarrow{l}}(h_{k+1}^{\overleftarrow{l}},Wh_{k+1}^{(l)}),
\end{equation}
\begin{equation}
h_k^{(l)}=[h_k^{\overrightarrow{l}}, h_k^{\overleftarrow{l}}],
\end{equation}
where $W$ is a weight parameter and $Wh_{k}^{(l)}$ is the hidden state for predicting future turns. If $K=1$, it means our model has no looking-ahead ability and it degrades to a naive goal-based sequence-to-sequence model.

\subsection{Decoding Module}

For generating the real system utterance, as seen in Figure~\ref{fig:model}, the green hidden states $\{Wh_k^{(l)}|k>0\}$ are combined through an attention based model~\cite{attentionlstm}. The formal denotation is:

\begin{equation}
e_{k}=tanh(W^{(a)}Wh_{k}^{(l)})],
\end{equation}
\begin{equation}
v_{k}=\frac{\exp{(e_k)}}{\sum_{k=1}^{K}\exp(e_k)},
\end{equation}
\begin{equation}
r=\sum_{k=1}^{K}{v_{k}}{h_{k}^{(l)}},
\end{equation}
where $W^{(a)}$ is the attention weight parameter and $r$ is the input representation for generating a new $u_{t+1}$ that is regarded as the real system utterance.

Given the hidden state $Wh_k^{(l)}$, the decoding module can also generate the corresponding utterance for learning the looking-ahead ability. We share the parameters of decoding with those in the encoding module, in order to reduce the computing cost~\cite{neuralconvmodel}. The token sequence in $u_{t+k}$ is generated from left to right by selecting the tokens with the maximum probability distribution through a language model learned by the following equation:
\begin{equation}
p_\theta(y_j^{(t+k)}|y_{1,2,...,j-1}^{(t+k)})\propto\exp(E^TWh_k^{(l)}).
\end{equation}

\subsection{Model Training}

To train the proposed model, we define a loss function to maximize three terms: (1) a language model for predicting tokens in language generation, (2) the probability distribution of predicting utterances of future dialogue turns, and (3) a binary classifier to predict if the dialogue will be complete or not. The final joint loss function is formally denoted as:
\begin{equation}
\begin{aligned}
L(\theta)=&-\underbrace{\sum_{u}\sum_i\log p_\theta(x_i|x_{1,...,i-1})}_{language~model~loss}\\&-\alpha\underbrace{\sum_{u,g}\sum_k\sum_i\log p_\theta(y_i^{(t+k)}|y_{1,...,i-1}^{(t+k)},u,g)}_{looking~ahead~prediction~loss}\\&-\beta\underbrace{\sum_c\log p(z_c|c,u_{t+1})}_{dialog~state~prediction~loss},
\end{aligned}
\end{equation}
where
\begin{equation}
u_{t+1}=\arg max_yp_\theta(y|r),
\end{equation}
\begin{equation}
\begin{aligned}
\log p(z_c|c,u_{t+1})&=z_c\log(g(c,u_{t+1}))\\&+(1-z_c)\log(1-g(c,u_{t+1})).
\end{aligned}
\end{equation}
$g(\cdot)$ is a sigmoid function and $z_c$ is the label of the dialogue that current user utterance $c$ belongs to, where 1 means the dialogue ends up with goals achieved while 0 means the goals are not achieved. The three terms are weighted with two hyper-parameters $\alpha$ and $\beta$. We adopt stochastic gradient descent method to minimize $L(\theta)$.

In the looking-ahead module, the hidden state $Wh_k^{(l)}$ is used to generate an utterance $y^{(t+k)}$, and is also used to calculate $h_{k+1}^{\overrightarrow{l}}$ and $h_{k-1}^{\overleftarrow{l}}$. We design an EM-like algorithm to optimize the loss function, as described in Algorithm 1. Line 3-4 optimize the language model, i.e. the first term of $L(\theta)$. Line 5-16 optimize the looking-ahead module, i.e. the second term, among which Line 7-14 are for E-step and Line 15-16 are for M-step. In E-step the language model is fixed for updating all the hidden states $h_k^{(l)}$ in looking-ahead module, and in M-step all the hidden states are fixed for updating the language model. Line 17-18 optimize the third term of $L(\theta)$, which is a binary classifier.

\begin{algorithm}\small
  \caption{Learning algorithm for $L(\theta)$}
  \SetKwInOut{Input}{input}\SetKwInOut{Output}{output}
  \Input{Dialogue utterances $U$, Agent goals $g$, Looking-ahead turns $K$}
  \Output{Agent model $\theta$}
  \nl Randomly initializing parameters\;
  \nl \For{$c\in{U}$, $g$ and historical utterances $\{u\}$}
  {
    \nl \For{$x_i\in{c}$}
    {
      \nl   Optimizing $p_\theta(x_i|x_{1,...,i-1})$\;
    }
    \nl $h_1^{\overrightarrow{l}}=[h^{(g)},h^{(u)},h^{(c)}]$\;
    \nl $u_{t+1}=\arg max_yp_\theta(y|r)$\;
    E-Step: Update $h_k^{(l)}$ with fixed language model\;
    \nl \For{$k=1:K$}{
      \nl   $u_{t+k}=\arg max_yp_\theta(y|h_k^{(l)})$\;
      \nl   $h_{k}^{\overrightarrow{l}}=[h_{k-1}^{\overrightarrow{l}}, Wh_{k-1}^{(l)}]$\;
    }
    \nl $h_K^{\overleftarrow{l}}=h_K^{\overrightarrow{l}}$\;
    \nl \For{$k=K-1:1$}
    {
      \nl   $h_{k}^{\overleftarrow{l}}=[h_{k+1}^{\overleftarrow{l}}, Wh_{k+1}^{(l)}]$\;
    }
    \nl \For{$k=1:K$}
    {
      \nl   $h_k^{(l)}=[h_k^{\overrightarrow{l}}, h_k^{\overleftarrow{l}}]$\;
    }
    M-Step: Update language model with fixed $h_k^{(l)}$\;
    \nl \For{$k=1:K$}
    {
      \nl   Optimizing $p_\theta(y_i^{(t+k)}|y_{1,...,i-1}^{(t+k)})$\;
    }
    \nl $u_{t+1}=\arg max_yp_\theta(y|r)$\;
    \nl Optimizing $p(z_c|c,u_{t+1})$\;
  }
  \nl \Return $\theta$\;
\end{algorithm}

\section{Experiments}


\subsection{Data Collection}

We use two datasets for two different scenarios to evaluate our model. Table~\ref{tab:stat} shows the statistics of two datasets.

\subsubsection{Dataset 1 - Object Division}

Dataset 1 contains crowd-sourced dialogues between humans collected from Amazon Mechanical Turk platform~\cite{deal}. The dataset is for \textit{object division task} and both sides have separate goals of each object's value. We use the textual data and transform their goals to yes-no questions as our binary vectors. The information of each dialogue session's final state, agree or disagree, is used for training the agent.

\subsubsection{Dataset 2 - Restaurant Reservation}

To the best of our knowledge, there is no other public dataset for goal-oriented dialogues where the two sides have different goals. To this end, we construct the Dataset 2 to testify the scalability of our model. The common scenario of restaurant table reservation is chosen.

In this dataset, the two agents are expected to have different goals and they dialogue with each other for looking for the intersection of their goals. We denote Agent A as the role of a customer and Agent B as the restaurant server side. At the beginning of each dialogue session, Agent A is given the available time slot, the number of people, and several other constraints (e.g. can sit at bar or not). All the constraints are regarded as its goals represented by a binary vector. Similarly, Agent B has itself constraints (e.g. whether bar tables are available or not), which are also treat as goals represented by a binary vector. We predefine a pool of `goals' and at the beginning of each dialogue session, the goals for two sides are randomly sampled separately from the pool. The two agents cannot see each other's goals and they dialogue through natural language until a final decision, agreement or disagreement, is reached. In summary, the objective of constructing this dataset is to see if our model can reach the intersection of the two agents' goals in a more efficient way.

To generate dialogues for Dataset 2, we resort to a rule-based method via AI planning search~\cite{planning,dataset}. Watson AI platform~\footnote{https://www.ibm.com/watson/ai-assistant/} is leveraged for natural language understanding by defining intents and entities with examples. A planner is designed for the dialogue manager by defining several states and actions. The goals are represented as part of the states, and the STRIPS algorithm is used to search the shortest path to goals at each turn and return the first planned action for generating the next response. Each action has several handcrafted utterances since the diversity of utterances is not our focus in this paper. Table~\ref{tab:sample} shows a sample dialogue.

\begin{table}\small
  \centering
  \begin{tabular}{|c|c|c|}
    \hline
    Metric & Dataset 1 & Dataset 2 \\
    \hline
    Number of Dialogues & 5,808 & 1,613 \\
    Average Turns per Dialogue & 6.6 & 6.3 \\
    Average Words per Turn & 7.6 & 8.9 \\
    Number of Words & 566,779 & 98,726 \\
    \% Goal Achieved & 80.1\% & 71.5\% \\
    \hline
  \end{tabular}
  \caption{Statistic on the two datasets.}
  \label{tab:stat}
\end{table}
\begin{table}\small
  \centering
  \begin{tabular}{|l|}
    \hline
    Alice: May I reserve a table for 6 people at 17 tomorrow? \\
    Bob: Sorry, we don't have a table at this point. \\
    Alice: Can we sit at the bar then? \\
    Bob: We don't have a bar in the restaurant. \\
    Alice: Can I have more expensive tables then? \\
    Bob: My apologies, we are required not to do that. \\
    Alice: In this case, can I reserve a bigger table? \\
    Bob: Yes, we have VIP rooms but more expensive. \\
    Alice: I want that. \\
    Bob: OK.\\
    Alice: Bye.\\
    \hline
  \end{tabular}
  \caption{Sample of Dataset 2.}
  \vspace{-2ex}
  \label{tab:sample}
\end{table}


\subsection{Training Sample Preparation}

For each dialogue session with $T$ turns, we re-organize the utterances into $T$ samples. For each turn $t=\{1,2,...,T\}$, we can get the current user utterance $c$, and a training sample is created with a historical utterance sequence $\{u_1,u_2,...,u_{t-1}\}$, and the goals $g$ are consistent with the same dialogue session. The future $K$ turns of utterances $\{u_{t+1},u_{t+2},...,u_{t+K}\}$ are used as the supervised information. In total, we get 38,333 and 10,162 samples including training set and test set for the two datasets respectively.

\subsection{Baselines}

Since our model is based on purely data-driven learning, we compare our model with the supervised counterparts. 
Our baselines include:

\begin{itemize}
  \item Seq2Seq(goal): This is a naive baseline by adapting the sequence-to-sequence model~\cite{seq2seq} and encoding goals, which removes the looking-ahead module and the supervised information of final state prediction from our model.
  \item Seq2Seq(goal+state): This is a baseline model by removing the looking-ahead module from our proposed model. The parameter $\alpha$ is set to zero.
  \item Seq2Seq(goal+look): This is a baseline model by removing the supervised information of final state prediction from our model. The parameter $\beta$ is set to zero.
  \item Seq2Seq(goal+look+state): This is our proposed model that includes all the modules and supervised information.
\end{itemize}

\subsection{Evaluation Criteria}

In a dialogue system, it could be treat as efficient if it obtains more final goal achievement with as few as possible dialogue turns. Thus we set two criteria for evaluating and comparing models adopted in our experiments: (1) the \textit{goal achievement ratio} that means the ratio of the number of goal achieved dialogue over the number of attempted dialogues), and (2) the \textit{average dialogue turns}.

\subsection{Evaluator}

Our experiments are to achieve goals through conversations, and it is difficult to directly adopt existing simulators~\cite{simulator1}. We refer to the work~\cite{simulator2} and fine-tune it to our task. For each dataset, a naive sequence-to-sequence model that encodes goals is regarded as the user simulator. We run 1000 times of dialogue sessions using the simulator.

Apart from using the simulator, we also invite humans to dialogue with the agents for 100 times each person for each dataset and we report the average results.

\begin{table*}\footnotesize
	\centering
	\begin{tabular}{|c|cc|cc|cc|cc|}
		\hline
		\multirow{3}{*}{Model} &
		\multicolumn{4}{c|}{Dataset 1} &
		\multicolumn{4}{c|}{Dataset 2} \\
		\cline{2-9} & \multicolumn{2}{c|}{\textit{vs. Simulator}} &
		\multicolumn{2}{c|}{\textit{vs. Human}} &
		\multicolumn{2}{c|}{\textit{vs. Simulator}} &
		\multicolumn{2}{c|}{\textit{vs. Human}} \\
		
		\cline{2-9} & \% Achieved & \# Turns & \% Achieved & \# Turns & \% Achieved & \# Turns & \% Achieved & \# Turns \\
		\hline
		Seq2Seq(goal) & 76.00 & 4.74 & 67.74 & 7.87 & 67.10 & 7.38 & 54.1 & 7.56 \\
		Seq2Seq(goal+state) & 79.41 & 4.74 & 70.97 & 6.35 & 67.37 & 7.42 & 58.1 & 8.04 \\
		Seq2Seq(goal+look) & 80.64 & 6.54 & 74.19 & 5.41 & 83.54 & \textbf{5.82} & 60.3 & \textbf{6.94} \\
		Seq2Seq(goal+look+state) & \textbf{85.07} & \textbf{4.10} & \textbf{77.42} & \textbf{5.02} & \textbf{83.58} & 6.36 & \textbf{61.2} & 7.30 \\
		\hline
	\end{tabular}
	\caption{Performance on two datasets against the user simulator and human.}
	\vspace{-3ex}
	\label{tab:result1}
\end{table*}

\begin{figure}
	\begin{centering}
		\subfloat{\begin{centering}
				\includegraphics[width=1.5in]{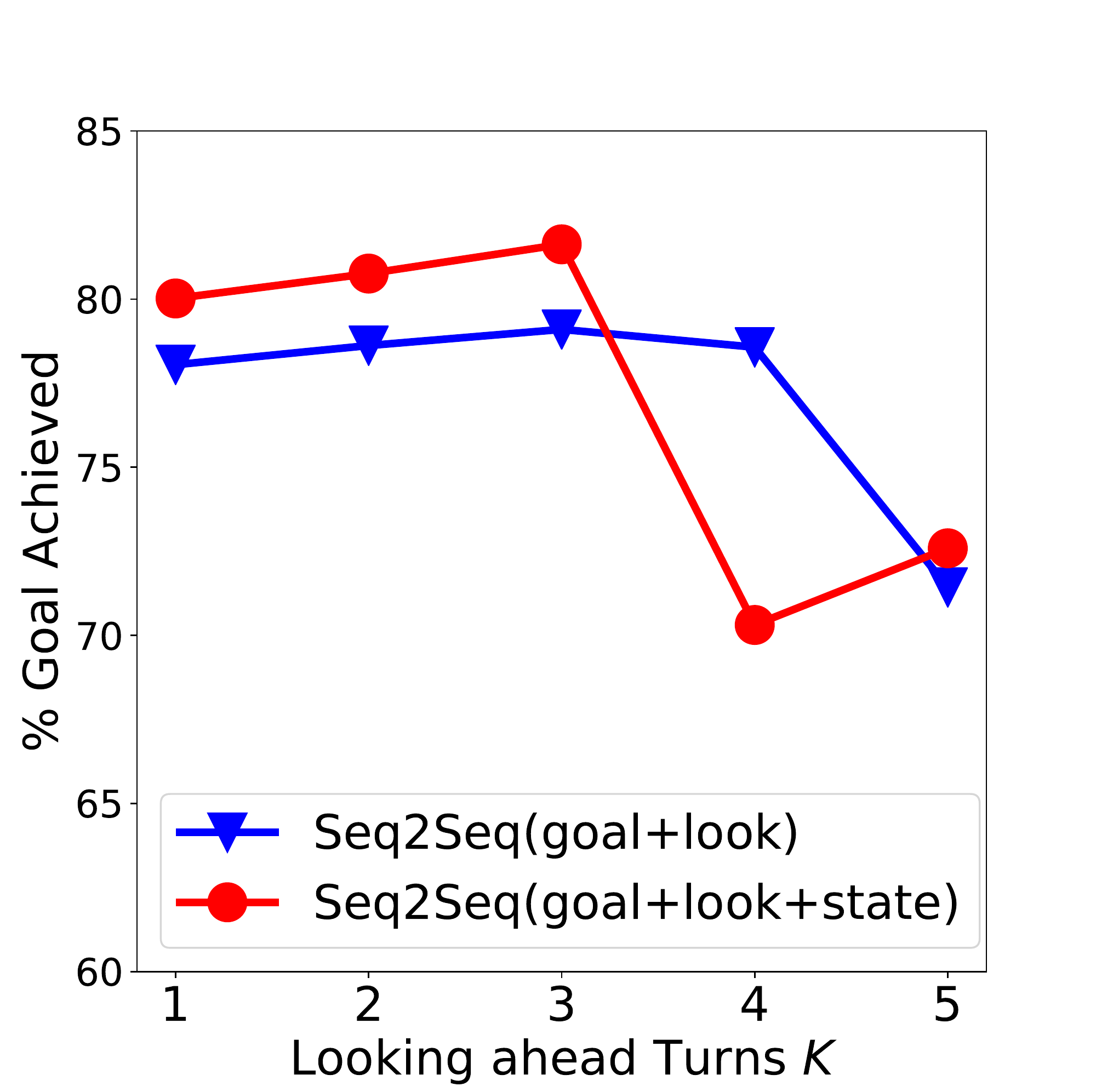}
				\par\end{centering}}
		\subfloat{\begin{centering}
				\includegraphics[width=1.5in]{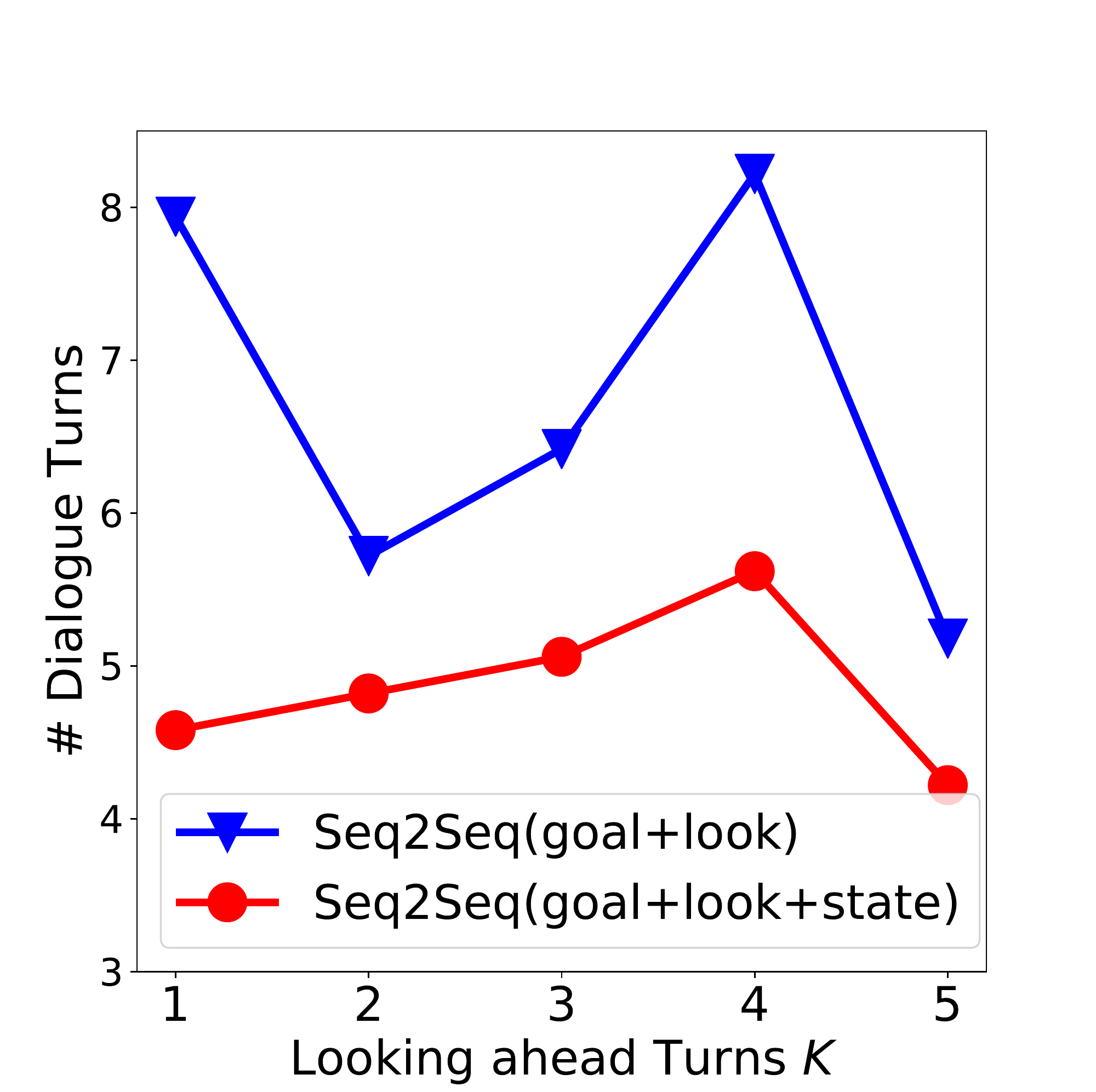}
				\par\end{centering}}
		\par\end{centering}
	\caption{vs. looking-ahead turns on Dataset 1}~\label{fig:stepdata1}
	\vspace{-4ex}
\end{figure}
\begin{figure}
	\begin{centering}
		\subfloat{\begin{centering}
				\includegraphics[width=1.5in]{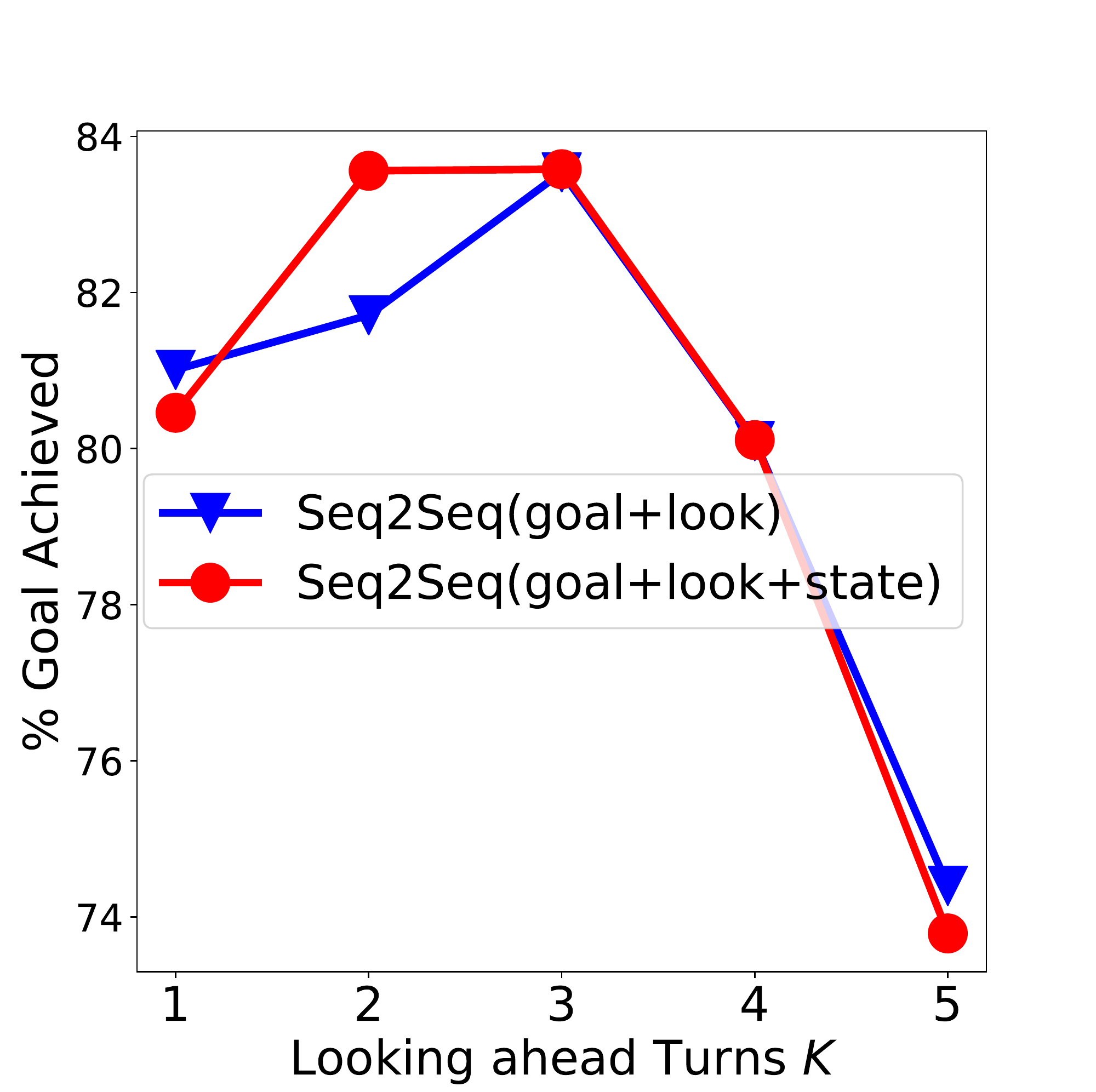}
				\par\end{centering}}
		\subfloat{\begin{centering}
				\includegraphics[width=1.5in]{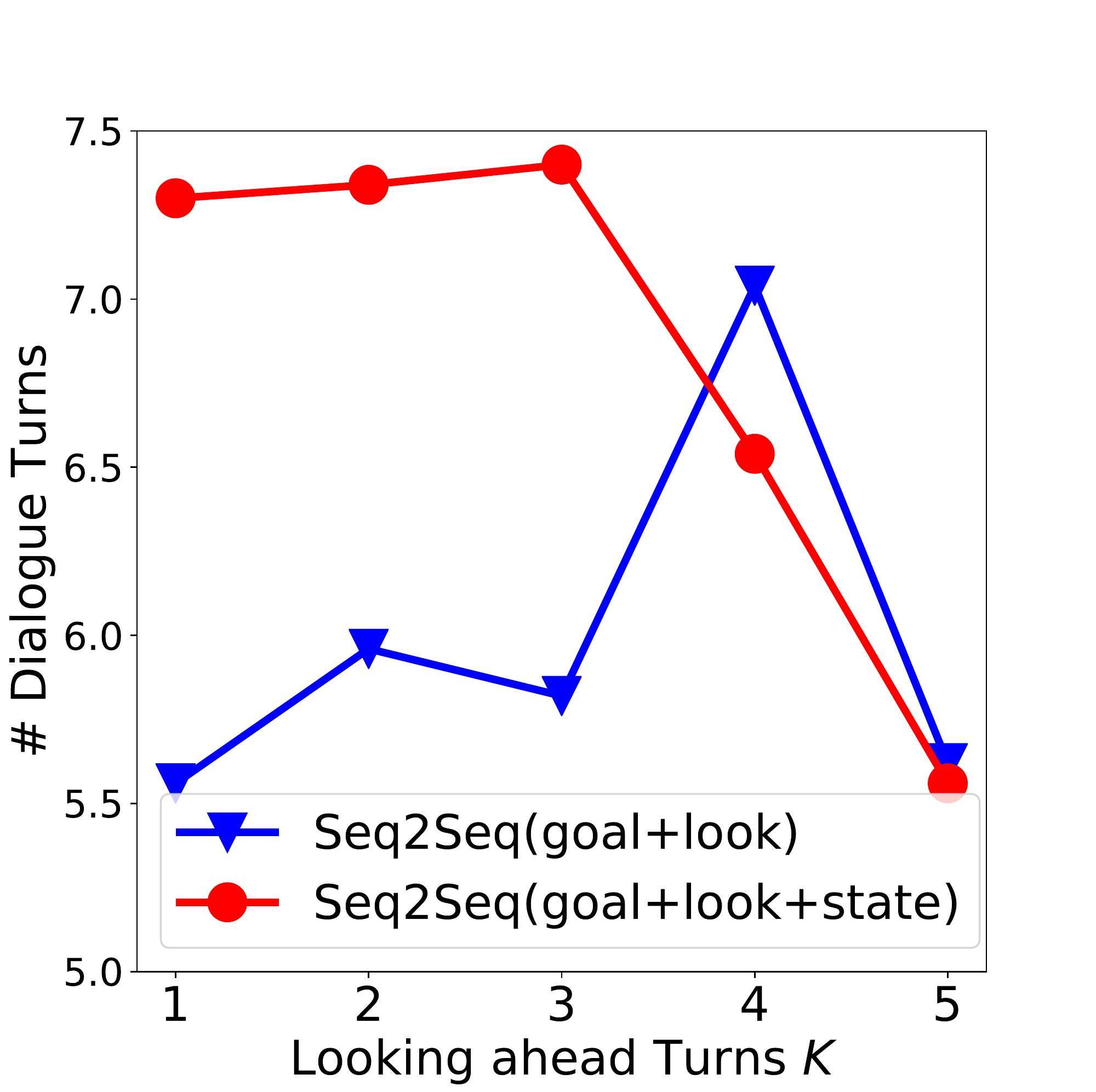}
				\par\end{centering}}
		\par\end{centering}
	\caption{vs. looking-ahead turns on Dataset 2}~\label{fig:stepdata2}
	\vspace{-4ex}
\end{figure}
\begin{figure}
	\begin{centering}
		\subfloat{\begin{centering}
				\includegraphics[width=1.5in]{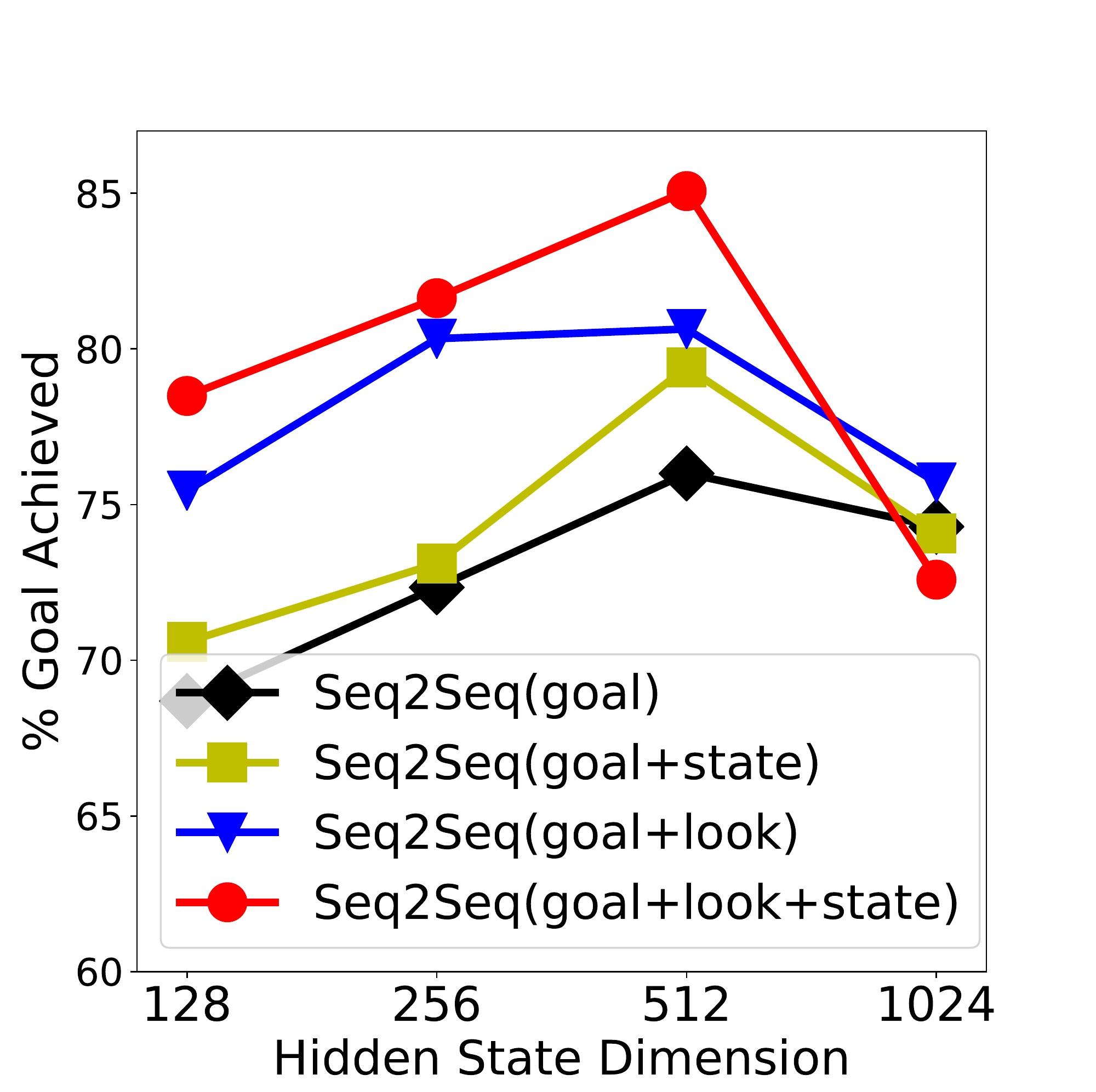}
				\par\end{centering}}
		\subfloat{\begin{centering}
				\includegraphics[width=1.5in]{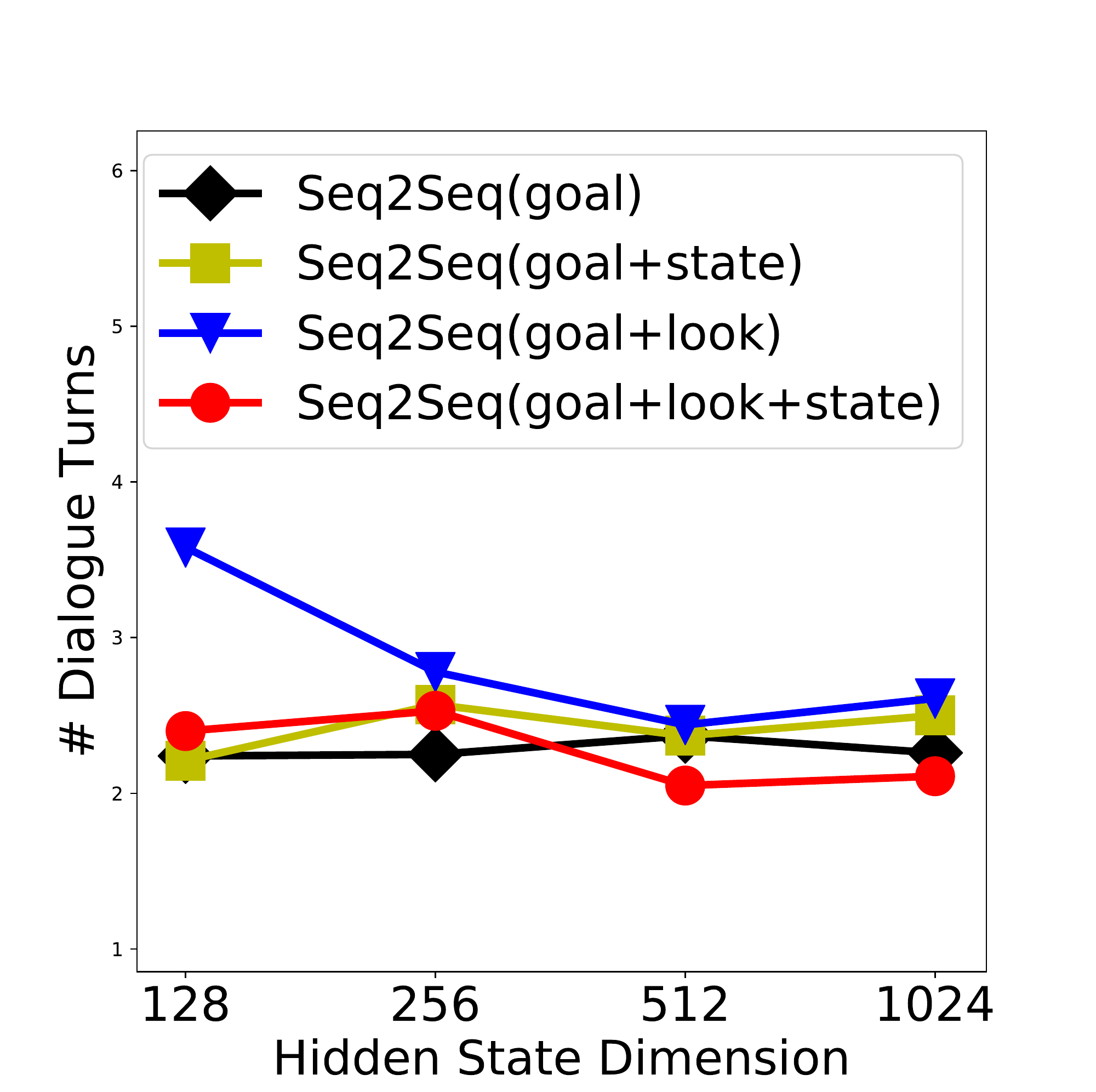}
				\par\end{centering}}
		\par\end{centering}
	\caption{vs. hidden state dimension on Dataset 1}~\label{fig:dimendata1}
	\vspace{-4ex}
\end{figure}

\begin{figure}
	\begin{centering}
		\subfloat{\begin{centering}
				\includegraphics[width=1.5in]{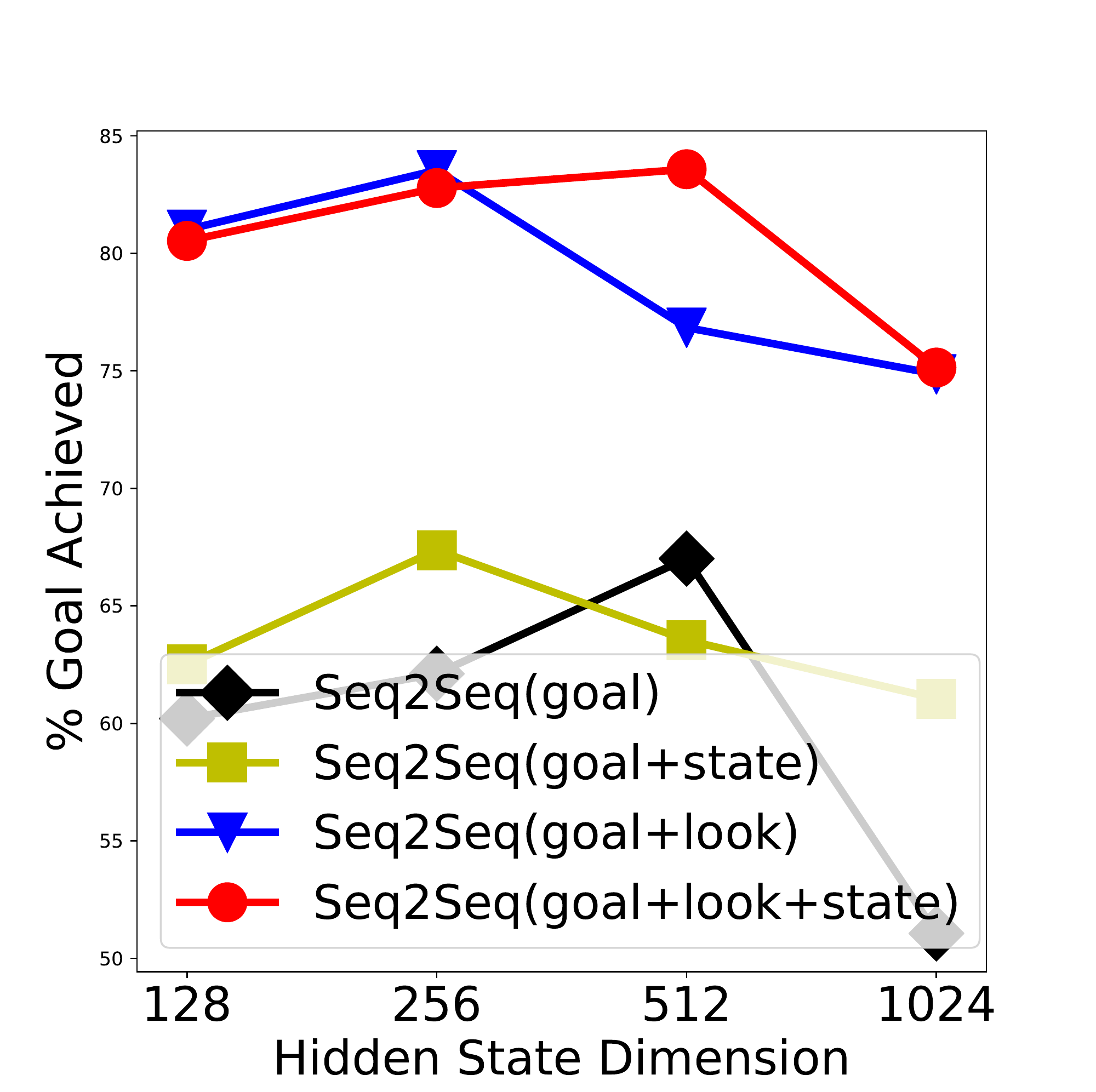}
				\par\end{centering}}
		\subfloat{\begin{centering}
				\includegraphics[width=1.5in]{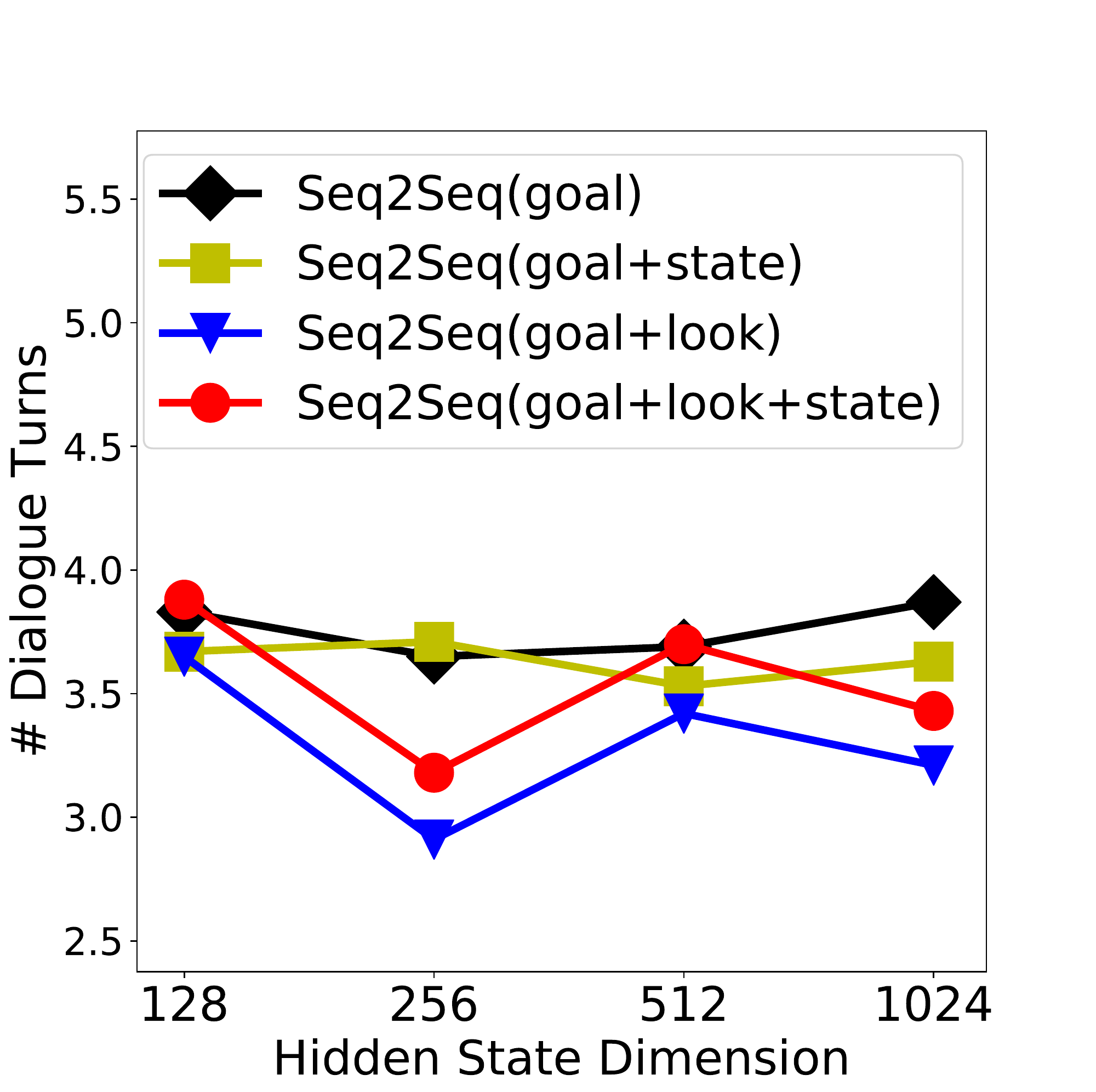}
				\par\end{centering}}
		\par\end{centering}
	\caption{vs. hidden state dimension on Dataset 2}~\label{fig:dimendata2}
	\vspace{-4ex}
\end{figure}

\subsection{Training Settings}

All the baselines are implemented by PyTorch. One-hot input tokens are embedded into a 64-dimensional space. The goals are encoded by $GRU^{(g)}$ with a hidden layer of size 64. The sizes of hidden states in input utterance encoder $GRU^{(u)}$, $GRU^{(c)}$ and looking-ahead module $GRU^{(l)}$, $h_k^{(l)}$, are all set to 256. A stochastic gradient descent method is employed to optimize the model with a mini-batch size of 32 for supervised learning, an initial learning rate of 1.0, momentum with $\mu=0.1$, and clipping gradients 0.5 in $L^2$ norm. The best model is chosen from the processing of training the model for 400 epochs. After that, the learning rate decays by a factor of 2 for every epoch. The initial hyper-parameters setting in the loss function (Equation (11)) is $\alpha=0.05$ and $\beta=1.0$. Words that appear in the training dataset for less than 5 times are replaced with the `unknown' ($\left<unk\right>$) token. A validation dataset is employed to choose the optimal hyper-parameters.


\subsection{Results and Analysis}

Table~\ref{tab:result1} shows the performance of baselines against user simulator and human on the two datasets. Both reveal that models that learn looking-ahead ability can achieve better performance and deliver more efficient dialogues in terms of both goal achievement ratio and dialogue turns. However, in the table, the dialogue turns of Seq2Seq(goal+look+state) are larger than those of Seq2Seq(goal+look), which may suggest that more dialogue turns lead to more achievement. In spite of this, the looking-ahead ability learned by our model is demonstrated to be effective on the two different scenarios. Moreover, the supervised information of final states (the third term of Equation (11)) is also proven effective in delivering more achievement, which can be seen from the second and last rows of Table~\ref{tab:result1}. Compared to the human evaluation, the results with the simulator generally are better. It is probable that human evaluators tend to be more rigorous and more turns are necessary to achieve goals.

\begin{table}\small
	\centering
	\begin{tabular}{|l|}
		\hline
		\textit{Seq2Seq(goal) Model:}\\
		\hline
		Alice: i just want the book \\
		Bob: no way i have the book and you can take others \\
		Alice: bye\\
		$<$Conversation end$>$ \\
		\hline
		\textit{Seq2Seq(goal+look+state) Model:}\\
		\hline
		Alice: i keep the book you can have the hat and two balls\\
		Bob: i will take the book and you can have others\\
		Alice: can we make a deal? \\
		Bob: ok deal\\
		Alice: okay \\
		Bob: bye\\
		$<$Conversation end$>$ \\
		\hline
	\end{tabular}
	\caption{Example that tends to achieve goals on Dataset 1, vs. simulator.}
	\label{tab:sampledata1}
	\vspace{-2ex}
\end{table}

We also evaluate the models' performance with different parameters. Figure~\ref{fig:stepdata1} and Figure~\ref{fig:stepdata2} show the results of using different looking-ahead turns $K$ in the looking-ahead module on Dataset 1 and Dataset 2 respectively. In most cases, Seq2Seq(goal+look+state) is better than Seq2Seq(goal+look). Setting the number to 3 is the recommended value, since our scenarios are relatively simple and it is enough for the agents to look ahead for 3 dialogue turns. The results are consistent with those in Table~\ref{tab:result1}.

The performance on different dimension sizes of hidden states used in all  the baselines is evaluated and the results are shown in Figure~\ref{fig:dimendata1} and Figure~\ref{fig:dimendata2}. We find that setting the size to 512 yields the optimal results by balancing the performance and the computing cost in most cases.

In addition, we take some case studies for dialogue examples between the agents trained by Seq2Seq(goal+look+state) (proposed model) and the simulator (Seq2Seq(goal)) for the two datasets. \textit{Alice} denotes the agent and \textit{Bob} denotes the simulator. Table~\ref{tab:sampledata1} is a case where our model with the looking-ahead ability (Seq2Seq(goal+look+state)) tends to achieve goals while consume more dialogue turns. Table~\ref{tab:sampledata2} gives an example where our model not only tends to achieve goals but also takes fewer dialogue turns. Based on our observation, the language models for natural language generation is learned well in spite of that confused utterances are generated sometimes.

\begin{table}\small
	\centering
	\begin{tabular}{|l|}
		\hline
		\textit{Seq2Seq(goal) Model:}\\
		\hline
		Alice: can you help me book a table for 2 people at 6pm? \\
		Bob: sorry, we dont have a table at this point \\
		Alice: can i reserve the seats at the bar instead? \\
		Bob: yes you can\\
		Alice: can i book a table for us\\
		Bob: sure i have written down your reservation \\
		Alice: ok \\
		Bob: bye \\
		$<$Conversation end$>$ \\
		\hline
		\textit{Seq2Seq(goal+look+state) Model:}\\
		\hline
		Alice: can you help me book a table for 2 people at 6pm? \\
		Bob: sorry we dont have a table at this point \\
		Alice: can i reserve the seats at the bar instead? \\
		Bob: sure i have written down your reservation \\
		Alice: bye \\
		$<$Conversation end$>$ \\
		\hline
	\end{tabular}
	\caption{Example that costs less dialogue turns under the same goals on Dataset 2, vs. simulator.}
	\label{tab:sampledata2}
	\vspace{-2ex}
\end{table}
\section{Conclusion}

In this paper, we propose an end-to-end model towards the problem of how to learn an efficient dialogue manager without taking too many manual efforts. We model the looking-ahead ability for foreseeing several turns and then the agent can make a decision of what to say that leads the conversation to achieve goals with as few as possible dialogue turns. Experiments on two datasets from different domains demonstrate that our model is efficient in terms of goal achievement ratio and average dialogue turns. Our method is also scalable and can reduce error propagation due to the nature of end-to-end learning.

For the future work, we expect to investigate whether other kinds of abilities, such as reasoning ability, can be modeled for agent towards the problem. In addition to the efficiency issue, the quality of natural language generation should also be paid attention in order to guarantee the quality of overall dialogue system.

\section*{Acknowledgments}
The work is partially supported by SFSMBRP (2018YFB1005100), BIGKE (No. 20160754021), NSFC (No. 61772076 and 61751201), NSFB (No. Z181100008918002), CETC (No. w-2018018) and OPBKLICDD (No. ICDD201901). We thank Tian Lan, Henda Xu and Jingyi Lu for experiment preparation. We also thank the three anonymous reviewers for their insightful comments.

\bibliography{acl2019}

\begin{thebibliography}{42}
\expandafter\ifx\csname natexlab\endcsname\relax\def\natexlab#1{#1}\fi

\bibitem[{Asher et~al.(2012)Asher, Lascarides, Lemon, Guhe, Rieser, Muller,
  Afantenos, Benamara, Vieu, Denis, Paul, Keizer, and Degr\'{e}mont}]{neg2}
Nicholas Asher, Alex Lascarides, Oliver Lemon, Markus Guhe, Verena Rieser,
  Philippe Muller, Stergos Afantenos, Farah Benamara, Laure Vieu, Pascal Denis,
  S.~Paul, S.~Keizer, and C.~Degr\'{e}mont. 2012.
\newblock Modelling strategic conversation: The stac project.
\newblock In \emph{SemDial}, page~27.

\bibitem[{Asri et~al.(2016)Asri, He, and Suleman}]{simulator1}
Layla~El Asri, Jing He, and Kaheer Suleman. 2016.
\newblock A sequence-to-sequence model for user simulation in spoken dialogue
  systems.
\newblock In \emph{INTERSPEECH}, pages 1151--1155.

\bibitem[{Bahdanau et~al.(2014)Bahdanau, Cho, and Bengio}]{gru1}
Dzmitry Bahdanau, Kyunghyun Cho, and Yoshua Bengio. 2014.
\newblock Neural machine translation by jointly learning to align and
  translate.
\newblock ArXiv preprint arXiv:1409.0473.

\bibitem[{Bordes et~al.(2017)Bordes, Boureau, and Weston}]{goal1}
Antoine Bordes, Y-Lan Boureau, and Jason Weston. 2017.
\newblock Learning end-to-end goal-oriented dialog.
\newblock In \emph{ICLR}.

\bibitem[{Chen et~al.(2017)Chen, Liu, Yin, and Tang}]{jd}
Hongshen Chen, Xiaorui Liu, Dawei Yin, and Jiliang Tang. 2017.
\newblock A survey on dialogue systems- recent advances and new frontiers.
\newblock \emph{ACM SIGKDD Explorations Newsletter}, 19(2):25--35.

\bibitem[{Cho et~al.(2014)Cho, van Merri\"{e}nboer, Bahdanau, and Bengio}]{gru}
Kyunghyun Cho, Bart van Merri\"{e}nboer, Dzmitry Bahdanau, and Yoshua Bengio.
  2014.
\newblock On the properties of neural machine translation: Encoder-decoder
  approaches.
\newblock In \emph{SSST-8}, pages 103--114.

\bibitem[{Dhingra et~al.(2017)Dhingra, Li, Li, Gao, Chen, Ahmed, and
  Deng}]{movie}
Bhuwan Dhingra, Lihong Li, Xiujun Li, Jianfeng Gao, Yun-Nung Chen, Faisal
  Ahmed, and Li~Deng. 2017.
\newblock Towards end-to-end reinforcement learning of dialogue agents for
  information access.
\newblock In \emph{ACL}, pages 484--495.

\bibitem[{Dodge et~al.(2015)Dodge, Gane, Zhang, Bordes, Chopra, Miller, Szlam,
  and Weston}]{evaluate2}
Jesse Dodge, Andreea Gane, Xiang Zhang, Antoine Bordes, Sumit Chopra, Alexander
  Miller, Arthur Szlam, and Jason Weston. 2015.
\newblock Evaluating prerequisite qualities for learning end-to-end dialog
  systems.
\newblock ArXiv preprint arXiv:1511.06931.

\bibitem[{Ghallab et~al.(2016)Ghallab, Nau, and Traverso}]{planning}
Malik Ghallab, Dana Nau, and Paolo Traverso. 2016.
\newblock \emph{Automated Planning and Acting}.
\newblock Cambridge University Press.

\bibitem[{Henderson et~al.(2014{\natexlab{a}})Henderson, Thomson, and
  Williams}]{challenge}
Matthew Henderson, Blaise Thomson, and Jason Williams. 2014{\natexlab{a}}.
\newblock The second dialog state tracking challenge.
\newblock In \emph{SIGDIAL}, pages 263--272.

\bibitem[{Henderson et~al.(2014{\natexlab{b}})Henderson, Thomson, and
  Williams}]{dstc3}
Matthew Henderson, Blaise Thomson, and Jason~D. Williams. 2014{\natexlab{b}}.
\newblock The third dialog state tracking challenge.
\newblock In \emph{SLT}, pages 324--329.

\bibitem[{Henderson et~al.(2014{\natexlab{c}})Henderson, Thomson, and
  Young}]{errorpro1}
Matthew Henderson, Blaise Thomson, and Steve Young. 2014{\natexlab{c}}.
\newblock Word-based dialog state tracking with recurrent neural networks.
\newblock In \emph{SIGDIAL}, pages 292--299.

\bibitem[{Jiang et~al.(2019)Jiang, Ma, Lu, Yu, Yu, and Li}]{dataset}
Zhuoxuan Jiang, Jie Ma, Jingyi Lu, Guangyuan Yu, Yipeng Yu, and Shaochun Li.
  2019.
\newblock A general planning-based framework for goal-driven conversation
  assistant.
\newblock In \emph{AAAI}, pages 9857--9858.

\bibitem[{Joshi et~al.(2017)Joshi, Mi, and Faltings}]{goal2}
Chaitanya~K. Joshi, Fei Mi, and Boi Faltings. 2017.
\newblock Personalization in goal-oriented dialog.
\newblock In \emph{NIPS}.

\bibitem[{Lewis et~al.(2017)Lewis, Yarats, Dauphin, Parikh, and Batra}]{deal}
Mike Lewis, Denis Yarats, Yann~N. Dauphin, Devi Parikh, and Dhruv Batra. 2017.
\newblock Deal or no deal? end-to-end learning for negotiation dialogues.
\newblock In \emph{EMNLP}, pages 2443--2453.

\bibitem[{Li et~al.(2016{\natexlab{a}})Li, Galley, Brockett, Gao, and
  Dolan}]{chat2}
Jiwei Li, Michel Galley, Chris Brockett, Jianfeng Gao, and Bill Dolan.
  2016{\natexlab{a}}.
\newblock A diversity-promoting objective function for neural conversation
  models.
\newblock In \emph{NAACL}, pages 110--119.

\bibitem[{Li et~al.(2017)Li, Chen, Li, Gao, and Celikyilmaz}]{IJCNLP}
Xiujun Li, Yun-Nung Chen, Lihong Li, Jianfeng Gao, and Asli Celikyilmaz. 2017.
\newblock End-to-end task-completion neural dialogue systems.
\newblock In \emph{IJCNLP}, pages 733--743.

\bibitem[{Li et~al.(2016{\natexlab{b}})Li, Lipton, Dhingra, Li, Gao, and
  Chen}]{simulator2}
Xiujun Li, Zachary~C Lipton, Bhuwan Dhingra, Lihong Li, Jianfeng Gao, and
  Yun-Nung Chen. 2016{\natexlab{b}}.
\newblock A user simulator for task-completion dialogues.
\newblock \emph{arXiv preprint arXiv:1612.05688}.

\bibitem[{Lipton et~al.(2018)Lipton, Li, Gao, Li, Ahmed, and Deng}]{BBQ}
Zachary Lipton, Xiujun Li, Jianfeng Gao, Lihong Li, Faisal Ahmed, and Li~Deng.
  2018.
\newblock Bbq-networks: Efficient exploration in deep reinforcement learning
  for task-oriented dialogue systems.
\newblock In \emph{AAAI}, pages 5237--5244.

\bibitem[{Liu and Lane(2017)}]{errorpro2}
Bing Liu and Ian Lane. 2017.
\newblock An end-to-end trainable neural network model with belief tracking for
  task-oriented dialog.
\newblock In \emph{INTERSPEECH}, pages 2506--2510.

\bibitem[{Liu et~al.(2018)Liu, T\"{u}r, Hakkani-T\"{u}r, Shah, and
  Heck}]{withhuman}
Bing Liu, Gokhan T\"{u}r, Dilek Hakkani-T\"{u}r, Pararth Shah, and Larry Heck.
  2018.
\newblock Dialogue learning with human teaching and feedback in end-to-end
  trainable task-oriented dialogue systems.
\newblock In \emph{NAACL}, pages 2060--2069.

\bibitem[{Luo et~al.(2019)Luo, Huang, Zeng, Nie, and Sun}]{goal3}
Liangchen Luo, Wenhao Huang, Qi~Zeng, Zaiqing Nie, and Xu~Sun. 2019.
\newblock Learning personalized end-to-end goal-oriented dialog.
\newblock In \emph{AAAI}.

\bibitem[{Mikolov et~al.(2013)Mikolov, Sutskever, Chen, Corrado, and
  Dean}]{word2vec}
Tomas Mikolov, Ilya Sutskever, Kai Chen, Gregory~S. Corrado, and Jeffrey Dean.
  2013.
\newblock Distributed representations of words and phrases and their
  compositionality.
\newblock In \emph{NIPS}, pages 3111--3119.

\bibitem[{Norvig and Russell(1995)}]{ai}
Peter Norvig and Stuart~J. Russell. 1995.
\newblock \emph{Artificial Intelligence: A Modern Approach}.
\newblock Prentice Hall.

\bibitem[{Peng et~al.(2018)Peng, Li, Gao, Liu, and Wong}]{DDQ}
Baolin Peng, Xiujun Li, Jianfeng Gao, Jingjing Liu, and Kam-Fai Wong. 2018.
\newblock Deep dyna-q: Integrating planning for task-completion dialogue policy
  learning.
\newblock In \emph{ACL}, pages 2182--2192.

\bibitem[{Peng et~al.(2017)Peng, Li, Li, Gao, Celikyilmaz, Lee, and
  Wong}]{hierarchical}
Baolin Peng, Xiujun Li, Lihong Li, Jianfeng Gao, Asli Celikyilmaz, Sungjin Lee,
  and Kam-Fai Wong. 2017.
\newblock Composite task-completion dialogue policy learning via hierarchical
  deep reinforcement learning.
\newblock In \emph{EMNLP}, pages 2231--2240.

\bibitem[{Rastogi et~al.(2018)Rastogi, Gupta, and Hakkani-Tur}]{multitask1}
Abhinav Rastogi, Raghav Gupta, and Dilek Hakkani-Tur. 2018.
\newblock Multi-task learning for joint language understanding and dialogue
  state tracking.
\newblock In \emph{SIGDIAL}, pages 376--384.

\bibitem[{Ritter et~al.(2011)Ritter, Cherry, and Dolan}]{chat1}
Alan Ritter, Colin Cherry, and William~B. Dolan. 2011.
\newblock Data-driven response generation in social media.
\newblock In \emph{EMNLP}, pages 583--593.

\bibitem[{Shah et~al.(2018)Shah, Hakkani-T\"{u}r, Liu, and
  T\"{u}r}]{bootstraping}
Pararth Shah, Dilek Hakkani-T\"{u}r, Bing Liu, and Gokhan T\"{u}r. 2018.
\newblock Bootstrapping a neural conversational agent with dialogue self-play,
  crowdsourcing and on-line reinforcement learning.
\newblock In \emph{NAACL}, pages 41--51.

\bibitem[{Stent et~al.(2004)Stent, Prasad, and Walker}]{plan1}
Amanda Stent, Rashmi Prasad, and Marilyn Walker. 2004.
\newblock Trainable sentence planning for complex information presentation in
  spoken dialog systems.
\newblock In \emph{ACL}, page~79.

\bibitem[{Su et~al.(2016)Su, Gasic, Mrksic, Rojas-Barahona, Ultes, Vandyke,
  Wen, and Young}]{continue}
Pei-Hao Su, Milica Gasic, Nikola Mrksic, Lina Rojas-Barahona, Stefan Ultes,
  David Vandyke, Tsung-Hsien Wen, and Steve Young. 2016.
\newblock Continuously learning neural dialogue management.
\newblock ArXiv preprint arXiv:1606.02689.

\bibitem[{Sutskever et~al.(2014)Sutskever, Vinyals, and Le}]{seq2seq}
Ilya Sutskever, Oriol Vinyals, and Quoc~V. Le. 2014.
\newblock Sequence to sequence learning with neural networks.
\newblock In \emph{NIPS}, pages 3104--3112.

\bibitem[{Vinyals and Le(2015)}]{neuralconvmodel}
Oriol Vinyals and Quoc Le. 2015.
\newblock A neural conversational model.
\newblock ArXiv preprint arXiv:1506.05869.

\bibitem[{Walker et~al.(2007)Walker, Stent, Mairesse, and Prasad}]{plan2}
Marilyn Walker, Amanda Stent, Fran\c{c}ois Mairesse, and Rashmi Prasad. 2007.
\newblock Individual and domain adaptation in sentence planning for dialogue.
\newblock \emph{Journal of Artificial Intelligence Research}, 30.

\bibitem[{Wang et~al.(2016)Wang, Huang, Zhao, and Zhu}]{attentionlstm}
Yequan Wang, Minlie Huang, Li~Zhao, and Xiaoyan Zhu. 2016.
\newblock Attention-based lstm for aspect-level sentiment classification.
\newblock In \emph{EMNLP}, pages 606--615.

\bibitem[{Wen et~al.(2017)Wen, Vandyke, Mrk\v{s}i\'{c}, Ga\v{s}i\'{c},
  Rojas-Barahona, Su, Ultes, and Young}]{EACL}
Tsung-Hsien Wen, David Vandyke, Nikola Mrk\v{s}i\'{c}, Milica Ga\v{s}i\'{c},
  Lina~M. Rojas-Barahona, Pei-Hao Su, Stefan Ultes, and Steve Young. 2017.
\newblock A network-based end-to-end trainable task-oriented dialogue system.
\newblock In \emph{EACL}, pages 438--449.

\bibitem[{Williams et~al.(2017)Williams, Asadi, and Zweig}]{hybrid}
Jason~D. Williams, Kavosh Asadi, and Geoffrey Zweig. 2017.
\newblock Hybrid code networks: practical and efficient end-to-end dialog
  control with supervised and reinforcement learning.
\newblock In \emph{ACL}, pages 665--677.

\bibitem[{Williams and Young(2007)}]{pomdp}
Jason~D. Williams and Steve Young. 2007.
\newblock Partially observable markov decision processes for spoken dialogue
  systems.
\newblock \emph{Computer Speech \& Language}, 21(2):393--422.

\bibitem[{Williams and Zweig(2016)}]{kefu}
Jason~D. Williams and Geoffrey Zweig. 2016.
\newblock End-to-end lstm-based dialog control optimized with supervised and
  reinforcement learning.
\newblock ArXiv preprint arXiv:1606.01269.

\bibitem[{Zhang et~al.(2018)Zhang, Cui, Wang, Zhu, Li, Zhou, and Liu}]{chat3}
Wei-Nan Zhang, Yiming Cui, Yifa Wang, Qingfu Zhu, Lingzhi Li, Lianqiang Zhou,
  and Ting Liu. 2018.
\newblock Context-sensitive generation of open-domain conversational responses.
\newblock In \emph{COLING}, pages 2437--2447.

\bibitem[{Zhang et~al.(2019)Zhang, Liao, Huang, Zhu, and Chua}]{retail}
Zheng Zhang, Lizi Liao, Minlie Huang, Xiaoyan Zhu, and Tat-Seng Chua. 2019.
\newblock Neural multimodal belief tracker with adaptive attention for dialogue
  systems.
\newblock In \emph{WWW}, pages 2401--2412.

\bibitem[{Zhao and Eskenazi(2016)}]{SIGDIAL}
Tiancheng Zhao and Maxine Eskenazi. 2016.
\newblock Towards end-to-end learning for dialog state tracking and management
  using deep reinforcement learning.
\newblock In \emph{SIGDIAL}, pages 1--10.

\end{thebibliography}
\bibliographystyle{acl_natbib}

\end{document}